\documentclass{article}

\usepackage{microtype}
\usepackage{graphicx}
\usepackage{subcaption}
\usepackage{booktabs}

\usepackage[accepted]{icml2026}

\usepackage{amsmath}
\usepackage{amssymb}
\usepackage{mathtools}
\usepackage{amsthm}
\usepackage{multirow}
\usepackage[table]{xcolor}
\usepackage{hyperref}
\usepackage{url}

\definecolor{codegreen}{rgb}{0,0.6,0}
\definecolor{codegray}{rgb}{0.5,0.5,0.5}
\definecolor{codepurple}{rgb}{0.58,0,0.82}
\definecolor{backcolour}{rgb}{0.95,0.95,0.92}

\definecolor{mydarkred}{rgb}{0.8,0.02,0.02}
\definecolor{mydarkorange}{rgb}{0.40,0.2,0.02}
\definecolor{mypurple}{RGB}{111,0,255}
\definecolor{myred}{rgb}{1.0,0.0,0.0}
\definecolor{mygold}{rgb}{0.75,0.6,0.12}
\definecolor{mydarkgray}{rgb}{0.66, 0.66, 0.66}
\definecolor{lightblue}{RGB}{220,235,245}
\definecolor{mygray}{gray}{0.9}

\usepackage{pifont} 
\newcommand{\greencheck}{\textcolor{green}{\ding{51}}}
\newcommand{\redcross}{\textcolor{red}{\ding{55}}}

\definecolor{oursblue}{HTML}{E8F6F4} 
\definecolor{myblue}{HTML}{71B7F7}
\definecolor{mypurple}{HTML}{C7AEF9}
\definecolor{myred}{HTML}{FF7C6E}
\usepackage{wrapfig}
\usepackage{caption}
\definecolor{lightblue}{rgb}{0.93,0.95,1.0}

\usepackage[capitalize,noabbrev]{cleveref}

\theoremstyle{plain}

\theoremstyle{definition}

\theoremstyle{remark}

\icmltitlerunning{Adaptive Visual Autoregressive Acceleration via Dual-Linkage Entropy Analysis}
\title{}
\begin{document}

\twocolumn[
  \icmltitle{Adaptive Visual Autoregressive Acceleration via Dual-Linkage Entropy Analysis}

  \begin{icmlauthorlist}
  \icmlauthor{Yu Zhang}{tongji}
  \icmlauthor{Jingyi Liu}{tongji}
  \icmlauthor{Feng Liu}{hit}
  \icmlauthor{Duoqian Miao}{tongji}
  \icmlauthor{Qi Zhang}{tongji}
  \icmlauthor{Kexue Fu}{sdas}
  \icmlauthor{Changwei Wang}{sdas}
  \icmlauthor{Longbing Cao}{mq}
\end{icmlauthorlist}

  \icmlaffiliation{tongji}{Tongji University}
\icmlaffiliation{hit}{Sagebot}
\icmlaffiliation{sdas}{Shandong Academy of Sciences}
\icmlaffiliation{mq}{Macquarie University}

\icmlcorrespondingauthor{Duoqian Miao}{dqmiao@tongji.edu.cn}

  \icmlkeywords{Machine Learning, ICML}

  \vskip 0.2in

    \begin{center}
{\includegraphics[width=\textwidth]{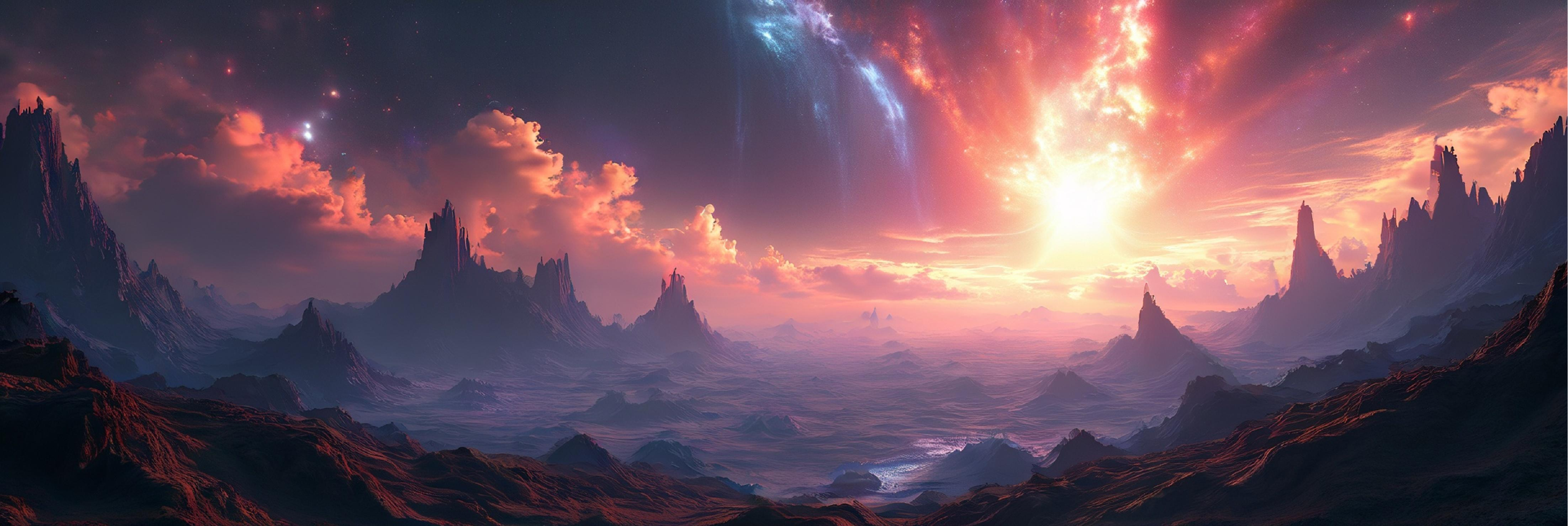}}
    \captionof{figure}{NOVA generates a 2K image in 3.2 seconds on a single NVIDIA RTX 3090 GPU without running out of GPU memory.}
    \label{fig:teaser}
  \end{center}
  
\vskip 0.2in

]

\printAffiliationsAndNotice{}

\begin{abstract}
Visual AutoRegressive modeling (VAR) suffers from substantial computational cost due to the massive token count involved. Failing to account for the continuous variation in modeling, existing VAR token reduction methods face three key limitations: heuristic stage partition, non-adaptive schedules, and limited acceleration scope, thereby leaving significant acceleration potential untapped. Since entropy variation intrinsically reflects the transition of predictive uncertainty, it offers a principled measure to capture continuous modeling variation. Therefore, we propose NOVA, a training-free token reduction acceleration framework for VAR models via entropy analysis. NOVA adaptively determines the acceleration activation scale during inference by online identifying the inflection point of scale entropy growth. Through scale-linkage and layer-linkage ratio adjustment, NOVA dynamically computes distinct token reduction ratios for each scale and layer, pruning low-entropy tokens while reusing the cache derived from the residuals at the prior scale to accelerate inference and maintain generation quality. Extensive experiments and analyses validate NOVA as a simple yet effective training-free acceleration framework. Code is \href{https://github.com/luokairo/NOVA}{\textcolor{magenta}{available}}.
\end{abstract}

\section{Introduction}

\begin{figure}[t]
     \centering
     \includegraphics[width=1\linewidth]{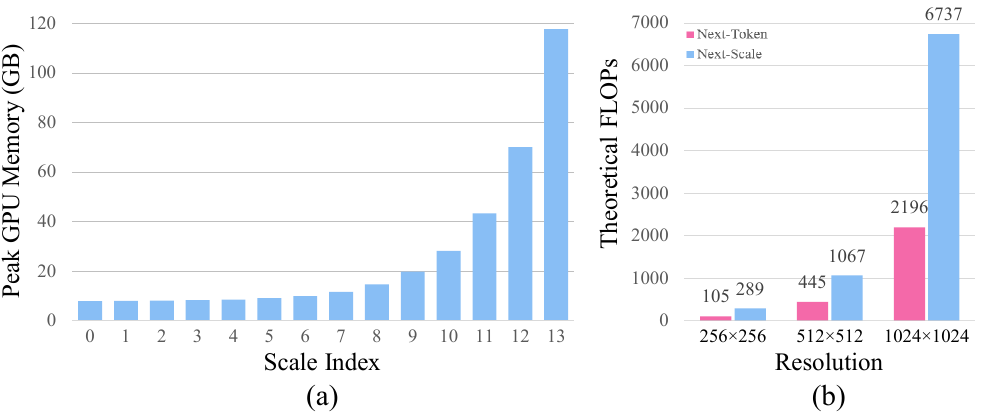}
     \vspace{-0.5cm}
     \caption{(a) Peak GPU memory consumption across different scales during 1024×1024 image generation in VAR. As the scale increases, the token sequence length grows rapidly, leading to quadratic growth in attention computation complexity. (b) Theoretical GFLOPs required by next-token and next-scale prediction models to generate images of various resolutions under uniform settings. Compared to the raster-scan next-token prediction, VAR's coarse-to-fine next-scale prediction additionally requires predicting feature maps at multiple previous scales before generating the final scale. The cumulative computation at these previous transition scales introduces extra computational cost to VAR.}
     \label{complexity}
     \vspace{-0.6cm}
 \end{figure}

Recently, Visual AutoRegressive modeling (VAR)~\cite{tian2024visual} based on next-scale prediction has demonstrated impressive breakthroughs in visual generation, achieving performance comparable to mainstream diffusion models~\cite{shen2025efficient,croitoru2023diffusion,cao2024survey}. However, VAR's performance improvement comes at the expense of a substantial computational cost. As illustrated in Figure~\ref{complexity}, its inefficiency mainly stems from the considerable computation complexity caused by the excessive token count, which further hinders its scalability to higher-resolution visual generation and integration into other downstream tasks.  

Towards direct and efficient application of VAR, training-free token reduction methods~\cite{guo2025fastvar,chen2025frequency,li2025stagevar} have emerged as promising acceleration solutions. These methods build on a common observation: VAR's substantial computational cost mainly occurs in the later stage. Consequently, they perform token reduction at this stage to accelerate inference. Although these methods achieve impressive results, they still face three key limitations: \textit{(1) Heuristic stage partition.} The later stage partition heavily relies on statistical estimation or empirical observation. 
The precision of this boundary is critical: premature token reduction risks quality degradation, while delayed token reduction limits acceleration gains. \textit{(2) Non-adaptive schedules.} All instances share the fixed stage boundary and token reduction ratio. However, different instance images vary enormously in complexity regarding semantics, composition, structure, lighting, and texture. The fixed stage partition and reduction ratio cannot guarantee optimality for every instance. \textit{(3) Limited acceleration scope.} Token reduction is performed only in the partitioned later stage, leaving potential computational redundancy in earlier stages unaddressed.

We attribute above limitations to a fundamental oversight: treating scales and layers as isolated units while neglecting the continuous variation in model predictions across scales and layers. Specifically, these methods rigidly partition the later stage and independently optimize the token reduction objectives for each scale or layer in the later stage. However, VAR inherently follows a coarse-to-fine transitional process that necessitates consideration of continuous modeling evolution across both scales and layers. Thus, we argue that it is essential to adopt a dual-linkage (scales and layers) adaptive token reduction method.

\begin{figure}[t]
    \centering
    \includegraphics[width=1\linewidth]{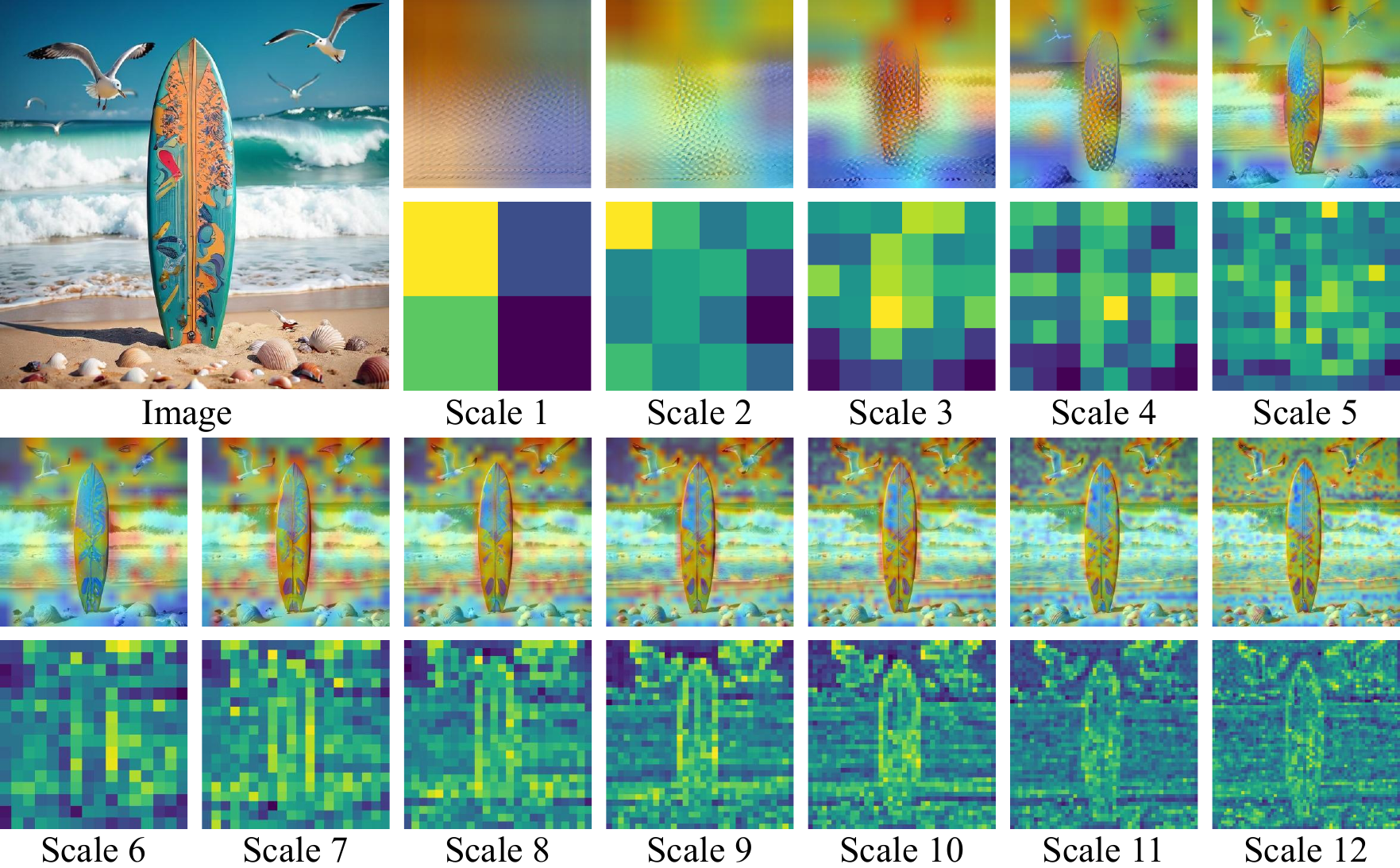}
    \vspace{-0.15cm}
    \caption{Entropy heatmaps of visualized images and token maps.}
    \label{heatmap}
    \vspace{-0.5cm}
\end{figure}

From the viewpoint of information theory, autoregressive generation is essentially a process of progressively reducing sequential uncertainty through successive observations. Each token generation represents a new observation. As a direct measure of uncertainty, entropy effectively reflects the change in predictive uncertainty throughout the process. A high-entropy token signifies a highly uncertain prediction with the newly observed information that may significantly impact subsequent distribution predictions, typically corresponding to larger potential for entropy reduction. Therefore, accurately modeling it offers substantial information gain\footnote{Information gain, a fundamental concept in information theory, quantifies the reduction in predictive uncertainty.} to further reduce sequential predictive uncertainty. Conversely, low-entropy tokens correspond to confident predictions for the current states with limited marginal impact on further narrowing distributions. Moreover, entropy variation allows us to analyze the uncertainty reduction process across scales or layers, thereby capturing the evolution of modeling dynamics. Meanwhile, as visualized in Figure~\ref{heatmap}, entropy heatmaps intuitively reveal the spatio-temporal diffusion characteristics of high-entropy regions: evolving from a globally concentrated pattern in early scales to a sparsely dispersed pattern in later scales. These analyses indicate that entropy can naturally can serve as a principled token selection criterion for our dual-linkage adaptive token reduction method.

In light of the above discussion, we propose NOVA, a training-free acceleration framework for VAR models. Guided by entropy analysis, NOVA performs a dual-linkage adaptive token reduction across both scales and layers. Specifically, at the scale level, it adaptively determines the scale for acceleration activation by online identifying the inflection point in scale entropy growth curve, and computes a distinct token reduction ratio for each acceleration scale by a scale-linkage function. At the layer level, NOVA further utilizes a layer-linkage ratio adjustment function to determine the specific reduction ratio for each layer and reuses the cache from residual interpolation of the same layer at the prior scale to improve speed and maintain performance.

Extensive experiments and analysis demonstrate the effectiveness of NOVA. On GenEval, NOVA achieved a 2.89× speedup for Infinity-2B with only 0.01\% performance loss, while surpassing comparable methods in both speed and performance. On ImageReward, NOVA not only reduced Infinity-8B's latency from 1.51s to 0.75s, but even achieved a preference score surpassing the original Infinity-8B. We further provide extensive visualizations to demonstrate the NOVA's excellent preservation of details and semantics.

\section{Related Work}
\paragraph{Next-Scale Visual AutoRegressive Generation.} 

By converting raster-scan order prediction to next-scale prediction, Visual AutoRegressive modeling (VAR)~\cite{tian2024visual} overcomes the limitations of next-pixel prediction and next-token prediction, which suffer from suboptimal performance due to disrupting the inherent spatial structure of images. Next-scale prediction and elevates generation quality to new heights. Some VAR-like variants, such as Infinity~\cite{han2025infinity}, HART~\cite{tang2024hart}, STAR~\cite{ma2024star} and VAR-CLIP~\cite{zhang2024var} extend VAR to text-to-image generation. Other improved VAR-like models~\cite{li2024controlvar,rajagopalan2025restorevar, xie2024litevar,wu2025nestedautoregressivemodels,jiao2025flexvarflexiblevisualautoregressive,qu2025visual,wang2025training} effectively perform various tasks.

\paragraph{Efficient Visual AutoRegressive Modeling.}

To address VAR's efficiency bottleneck, numerous efforts~\cite{huang2025spectralar,he2025neighboring,kumar2025scale,zhang2025markovian,zhang2025actvar} have been made. However, all these methods necessitate major architectural modifications~\cite{ren2024m,chen2025collaborative,vincenti2025dynamic} or extensive retraining~\cite{tang2024hart,kumbong2025hmar}, making them difficult to directly obtain and deploy. Training-free methods, such as FreqExit/SkipVAR's early-exit inference or decision-based acceleration~\cite{li2025skipvar,lifreqexit}, ScaleKV/HACK's KV-Cache optimization~\cite{li2025memory,qin2025head}, and LiteVAR's efficient attention and low-bit quantization~\cite{xie2024litevar}, achieve good speedups but face some challenges, such as generation quality degradation due to skipping entire layers or blocks instead of individual tokens and suboptimal performance due to input-independent static acceleration. In contrast, training-free token reduction methods~\cite{guo2025fastvar,chen2025frequency,li2025stagevar,kou2026pu} have led the development of efficient VAR due to their strong generalizability and flexibility. However, due to the lack of adaptivity and dynamics, their acceleration potential remains unleashed.

\paragraph{Entropy-Guided Deep Learning.} 
Entropy, as a measure of uncertainty and information distribution, is inherently linked to probabilistic modeling. Consequently, entropy-guided methods have been widely adopted in deep learning. Studies~\cite{peer2022improving, meni2024entropy, dubey2017regularizing} utilize entropy-based regularization to improve robustness and  stability. SEGA~\cite{wu2023sega}, MGEDE~\cite{yang2023minimum} and SEBot~\cite{yang2024sebot} enhance graph representation learning from an entropy perspective. 
EGGesture~\cite{xiao2024eggesture} and EDRL~\cite{wang2023edrl} propose entropy-guided methods for VQ-VAE and disentangled representation learning, respectively. 
Similar entropy principles also benefit label distribution learning (\cite{koumll, koubldl, kou2025nips, koutnnls, kou2026fedharmony, wu2026trustworthyfederatedlabeldistribution}).
Meanwhile, entropy is also employed to improve efficiency, such as accelerating robotic visuomotor policies~\cite{guo2025demospeedup}, video diffusion models~\cite{li2025efficient}, and diffusion language models~\cite{ben2025accelerated}. However, its application to visual autoregressive generation almost leaves blanks.

\begin{figure*}[t]
    \includegraphics[width=1\linewidth]{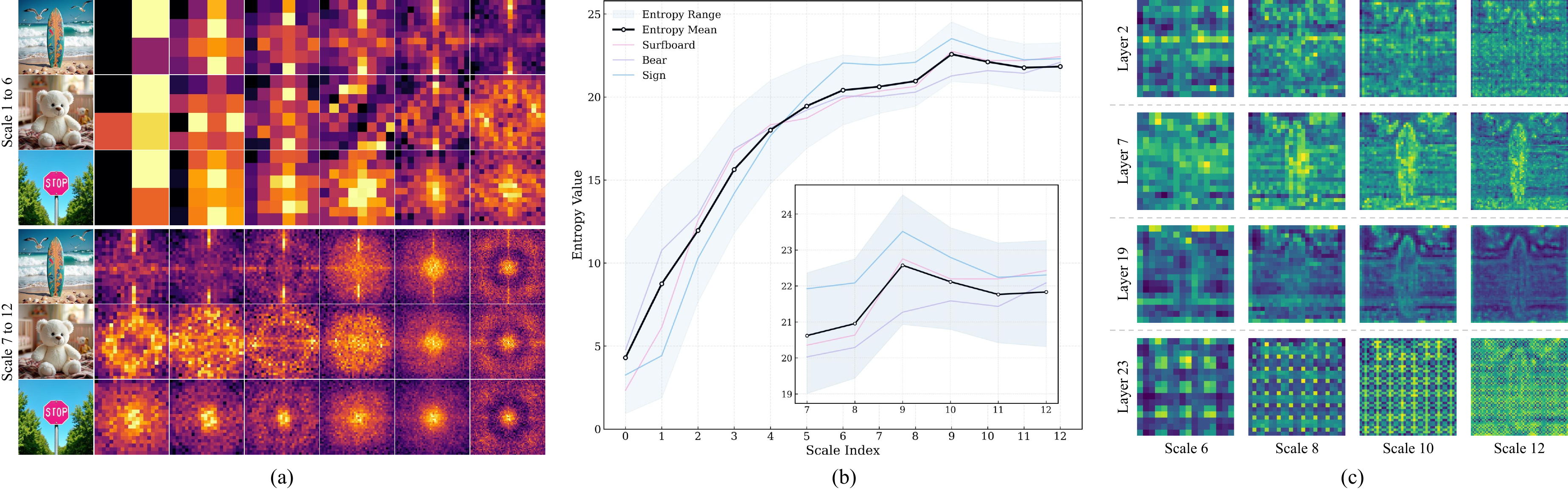}
    \caption{(a) Frequency spectrum of various instances at different scales during VAR generation. (b) Various entropy values across different scales during the inference process of Infinity-2B: We selected 600 prompts from GenEval and rewrote them. For each rewritten prompt, we generated four images using different random seeds. (c) Entropy heatmaps of token maps across different Transformer layers. Cross-layer spatial pattern variations: Shallow layers exhibit high-entropy regions resembling local noise or scattered dots, certain middle layers show high-entropy regions with clear subject contours, while deeper layers display regular periodic grid-like patterns in high-entropy regions. Cross-layer overall entropy variations: For example, token map at the 19-th layer has relatively low overall entropy.}
    \label{analysis}
\end{figure*}

\section{Methodology}

\subsection{Preliminary}
VAR~\cite{tian2024visual} reformulates autoregressive modeling from next-token prediction to next-scale residual feature prediction, generating images in a coarse-to-fine manner. Formally, the modeling process involves $T$ multi-scale token maps $\{R_1, R_2,\cdots,R_T\}=\{R_t\}_{t=1}^T$, where $R_t \in [V]^{h_t \times w_t}$ and $(h_t,w_t)$ denotes the grid size at scale $t$. At the $t$-th scale, VAR predicts the token map $R_t$ conditioned on all previous scales $R_{<t}=\{R_1,R_2, \cdots,R_{t-1}\}$. The autoregressive likelihood is formulated as follow:
\begin{equation}
    p(R_1, R_2,\cdots,R_T)
    ~=~
    \prod_{t=1}^{T}
    p(R_t \mid \langle\mathrm{sos}\rangle, R_{<t}),
    \label{eq:var_prob}
\end{equation}
where $\langle\mathrm{sos}\rangle$ denotes the start conditional embedding. 

\subsection{Analysis}

\paragraph{Theoretical Foundation for Entropy-Guided Token Reduction.}
In information theory, given a conditioning variable $Z$, the conditional mutual information between random variables $X$ and $Y$ satisfies
\begin{equation}\label{1}
I(X;Y\mid Z) \le H(X\mid Z),
\end{equation}
where $H(X | Z)$ denotes the conditional entropy of $X$ given $Z$, and conditional mutual information $I(X;Y| Z)=H(Y|Z)-H(Y|X,Z)$ measures information gain, i.e., the reduction in uncertainty of $Y$ after observing $X$ given $Z$.

VAR treats scales as autoregressive steps. For $t$-th scale, let $N_t=h_t w_t$ and denote the token at $R_t$ spatial index $i\in\{1,\dots,N_t\}$ as $r_{t,i}\in [V]$. Therefore, the upper bound of subsequent uncertainty at the $t$-th scale is determined by the sum of conditional entropies of all tokens in $R_t$: 
\begin{equation}\label{2}
H(R_t \mid R_{<t}) \le \sum_{i=1}^{N_t} H(r_{t,i} \mid R_{<t}) .
\end{equation}
Combining Equations~\ref{1} and~\ref{2}, it follows that maintaining highest-entropy tokens ensures the upper bound of uncertainty remains controllable while maximizing potential information gain to reduce future uncertainty, thereby facilitating the autoregressive generation process. Meanwhile, pruning the lowest-entropy tokens has limited impact on potential information gain but effectively reduces attention computation cost, thus providing theoretical foundation for our entropy-guided token reduction method.

\paragraph{Scale-Level Generation Process Analysis.}
Existing methods~\cite{guo2025fastvar,chen2025frequency,li2025stagevar} employ frequency to heuristically partition the later stage and perform token reduction within it. As shown in Figure~\ref{analysis} (a), we analyze the frequency spectrum evolution of token maps at different scales during generation across the three instances. Although a general coarse-to-fine process and certain stage-specific characteristics are observable, the distribution, content, and diffusion evolution of frequency components vary significantly across instances, making it difficult to adaptively and accurately partition the later stage through frequency analysis. Furthermore, some instances exhibit high-frequency patterns even in the middle scales, indicating more opportunities for token reduction prior to the later stage. While these works are highly novel and interesting and have achieved impressive results, we realize that relying solely on frequency, the physical property of images, makes it difficult to precisely capture the modeling process. Only by addressing the limitations of heuristic stage partition, non-adaptive schedules, and limited acceleration scope from the intrinsic properties of VAR generation, its acceleration potential can be further unleashed.

We calculate the predictive entropy for all instances across scales. Figure~\ref{analysis} (b) illustrates the overall entropy range and mean, while highlighting the entropy variations for the three aforementioned  instances: surfboard, bear, and sign. The subfigure reveals that the entropy of three instances fluctuates and diverges from one another at the larger scales, significantly deviating from the entropy mean. This suggests that even from an entropy perspective, it remains challenging to accurately and adaptively partition the later stage. Unlike previous works, this frees us from the focus on later stage partition. Nevertheless, the entropy range, the entropy mean, and the three instances all exhibit a consistent two-stage dynamic: entropy grows rapidly at early scales, followed by a phase of relatively slow growth or unstable fluctuation. This aligns with previous findings~\cite{li2025stagevar,chen2025tts,chen2025collaborative} that VAR models construct concepts and objects at early scales. This motivates  leveraging the entropy growth inflection point to adaptively activate acceleration, avoiding quality degradation from premature reduction and limited acceleration from delayed reduction.

\paragraph{Layer-Level Generation Process Analysis.} 
We further analyze the entropy characteristics of token maps across different Transformer layers at the same scale. Figure~\ref{analysis} (c)  reveals significant layer-level entropy heterogeneity. First, different layers exhibit various spatial distributions of high-entropy regions, indicating their distinct spatial modeling patterns. Second, the overall (average) entropy at the same scale also varies across layers. These variations suggest a uniform reduction ratio across all layers is suboptimal. Instead, token reduction at the same scale should be layer-adaptive, allocating more computation to high-entropy layers while reducing more tokens in low-entropy layers.

\subsection{NOVA}
\begin{figure}[t]
    \centering
    \includegraphics[width=1\linewidth]{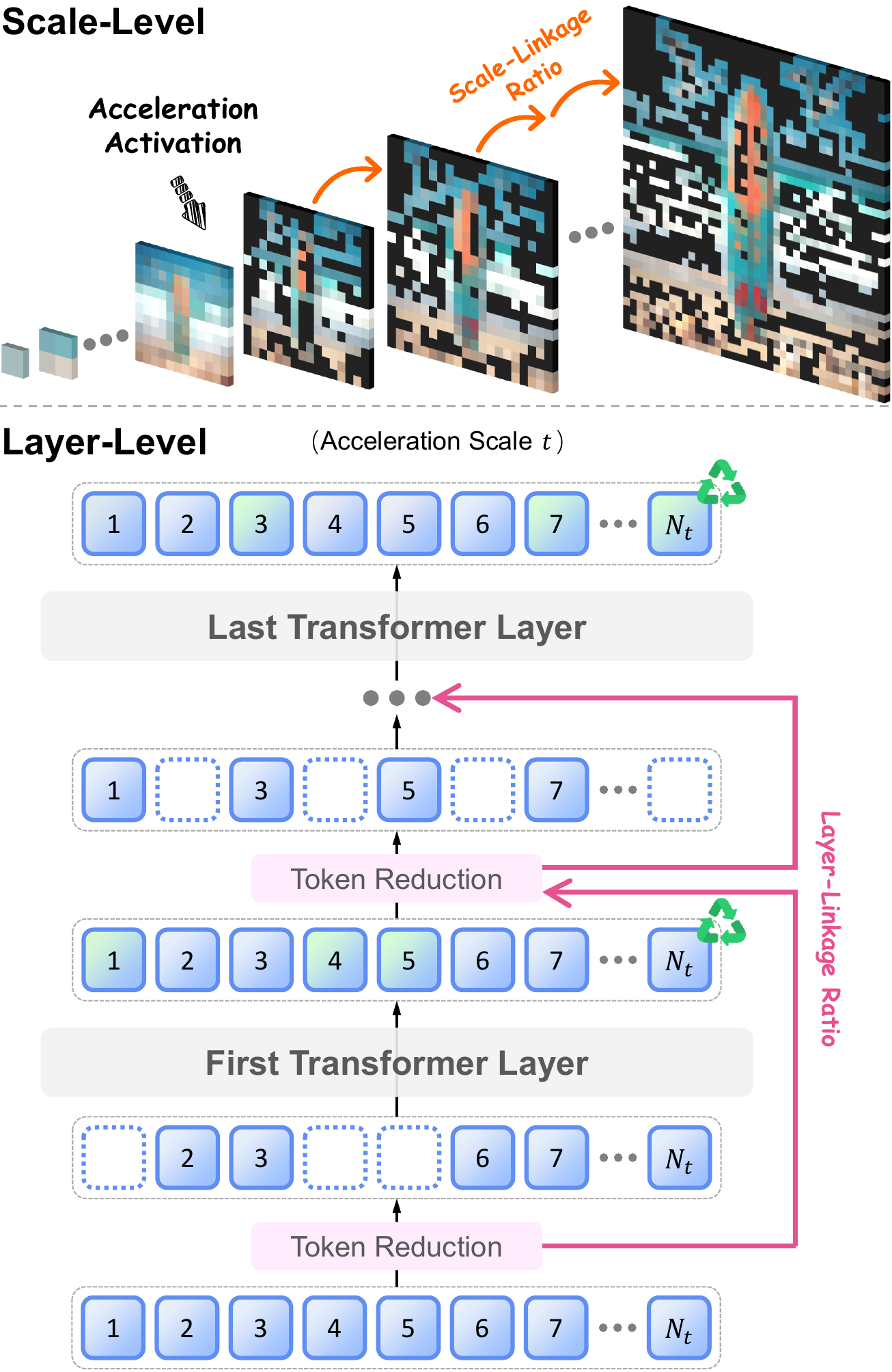}
    \caption{NOVA framework at both scale-level and layer-level.}
    \label{framework}
\end{figure}

Figure~\ref{framework} illustrates the NOVA framework at both scale-level and layer-level, which consists of three key components: Adaptive Acceleration Activation, Dual-Linkage Acceleration and Residual Cache Reuse.

\paragraph{Adaptive Acceleration Activation.}
At the scale-level, NOVA first determines the acceleration activation scale by online identifying the inflection point of the entropy growth across scales. For the token map $R_t$ at the $t$-th scale, the predictive entropy of the $i$-th token $r_{t,i}\in [V]$ is defined as:
\begin{equation}
    \mathcal{H}_{t,i}
    =
    H\!\left(r_{t,i}\mid R_{<t}\right)
    =
    -\sum_{v\in[V]} p_{t,i}(v)\log p_{t,i}(v),
\end{equation}
where $p_{t,i}(\cdot)$ is the predicted class distribution for token $r_{t,i}$.

Since VAR performs parallel prediction for $N_t$ tokens within the $t$-th scale, we adopt a factorized conditional likelihood within the token map $R_t$:
\begin{equation}
    p(R_t \mid R_{<t})
    =
    \prod_{i=1}^{N_t} p(r_{t,i}\mid R_{<t}).
\end{equation}
Therefore, the joint predictive entropy of the token map $R_t$ decomposes additively into token predictive entropy:
\begin{equation}
    H(R_t \mid R_{<t})
    =
    \sum_{i=1}^{N_t} H(r_{t,i}\mid R_{<t})
    =
    \sum_{i=1}^{N_t} \mathcal{H}_{t,i}.
\end{equation}
We aggregate the predictive entropies of all tokens in the token map $R_t$ and calculate their mean to obtain the $t$-th scale overall predictive entropy $\bar{\mathcal{H}}_{t}$:
\begin{equation}
    \bar{\mathcal{H}}_t
    \triangleq
    \frac{1}{N_t} H(R_t \mid R_{<t})
    =
    \frac{1}{N_t}\sum_{i=1}^{N_t}\mathcal{H}_{t,i}.
    \label{eq:scale_entropy_mean}
\end{equation}
In our earlier analysis, we found that the scale overall entropy $\bar{\mathcal{H}}_t$ exhibits a consistent two-stage dynamic as the scale increases. Therefore, using $\bar{\mathcal{H}}_t$ as a criterion, we online detect the decay of the entropy growth rate to identify the earliest scale where the growth rate decays and remains small as the acceleration activation scale. We define the discrete entropy growth rate $g_t$ at scale $t$ as:
\begin{equation}
    g_t
    \triangleq
    \bar{\mathcal{H}}_t - \bar{\mathcal{H}}_{t-1},
    \quad t = 2,\dots,T.
\end{equation}
As a training-free acceleration framework, NOVA cannot access the overall entropy of all future scales at the beginning of inference. Therefore, we preserve the first $t_{\mathrm{est}}$ scales without token reduction to estimate the early entropy growth baseline for the current instance:
\begin{equation}
    \eta
    \triangleq
    \frac{1}{t_{\mathrm{est}}-1}
    \sum_{t=2}^{t_{\mathrm{est}}} g_t,
    \label{eq:baseline_growth}
\end{equation}
Considering the volatile fluctuations in $\bar{\mathcal{H}}_t$, we perform a moving average $\tilde g_t$ on two adjacent discrete entropy growth rates $g_t$ obtained after the early entropy growth baseline estimation to mitigate misjudgments caused by oscillations: $\tilde g_t = (g_t+g_{t-1})/2$, where $t \ge t_{\mathrm{est}}+1$.

We identify the earliest scale $t^\star$ when the smoothed growth rate drops below a fraction $\alpha\in(0,1)$ of the early entropy growth baseline $\eta$:
\begin{equation}
    t^\star
    \triangleq
    \min\left\{
    t ~\big|~
    t \ge t_{\mathrm{est}}+1,\;
    \tilde g_t \le \alpha\, \eta
    \right\}.
    \label{eq:stabilization_point}
\end{equation}
The scale $t^\star$ is regarded as the inflection point of the scale overall entropy growth. Consequently, NOVA activates acceleration at the scale $t^\star +1$.

\paragraph{Dual-Linkage Acceleration.}
The core of dual-linkage acceleration lies in dynamically adjusting and determining the distinct token reduction ratio for each scale or layer based on entropy variations of token maps. We design scale-linkage and layer-linkage ratio functions to realize. 

For scale-linkage ratio function, we first assign each scale a base reduction ratio that progressively increases as generation proceeds. We then refine this base ratio by entropy growth rate $g_t$: if uncertainty is still changing rapidly, we reduce fewer tokens; if it trends toward stability, we reduce more. Thus, the token reduction ratio at the $t$-th scale is:
\begin{equation}
    Ratio_t = \sigma \left( \frac{t - (t^\star +1)}{\tau} \right) - \lambda \cdot \tanh(g_t),
    \label{eq:dual}
\end{equation}
where $\sigma(\cdot)$ is the sigmoid function, $\lambda$ is a scaling factor.

\begin{table*}[t]
    \centering
    \small
    \caption{Quantitative comparisons of perceptual quality on the GenEval and DPG-Bench benchmarks.}
    \vspace{-0.2cm}
    \label{tab:comparison}
    \setlength{\tabcolsep}{5pt}
    \renewcommand{\arraystretch}{1.05}
    
    \begin{tabular}{l cc | ccccc | ccc}
        \toprule
        \multirow{2}{*}{\textbf{Methods}} & \multirow{2}{*}{\textbf{Speed$\uparrow$}} & \multirow{2}{*}{\textbf{Latency$\downarrow$}} & \multirow{2}{*}{\textbf{Param$\downarrow$}} & \multicolumn{4}{c|}{\textbf{GenEval$\uparrow$}} & \multicolumn{3}{c}{\textbf{DPG-Bench$\uparrow$}} \\
        \cmidrule(lr){5-8} \cmidrule(lr){9-11}
        & & & & Two Obj. & Position & Color Attri. & \textbf{Overall} & Global & Relation & \textbf{Overall} \\
        \midrule
        HART & 1.00 $\times$ & 1.12s & 0.7B & 0.62 & 0.13 & 0.18 & 0.51 & 87.10 & 92.95 & 80.89 \\
        +FastVAR & 1.30 $\times$ & 0.86s & 0.7B & 0.57 & 0.16 & 0.19 & 0.50 & 86.21 & 92.48 & 80.71 \\
        +SparseVAR & 1.38 $\times$ & 0.81s & 0.7B & 0.60 & 0.12 & 0.21 & 0.50 & 86.63 & 92.54 & 80.70 \\
        \rowcolor{lightblue}
        +NOVA & 1.62 $\times$ & 0.69s & 0.7B & 0.60 & 0.13 & 0.17 & 0.51 & 86.88 & 92.53 & 80.84 \\
        \midrule
        Infinity-2B & 1.00 $\times$ & 2.43s & 2.0B & 0.85 & 0.45 & 0.54 & 0.73 & 85.10 & 92.37 & 83.12 \\
        +FastVAR & 2.45 $\times$ & 0.99s & 2.0B & 0.81 & 0.37 & 0.52 & 0.70 & 84.41 & 92.76 & 82.86 \\
        +SparseVAR & 1.66 $\times$ & 1.46s & 2.0B & 0.84 & 0.42 & 0.54 & 0.72 & 84.11 & 92.70 & 82.56 \\
        +SkipVAR & 2.23 $\times$ & 1.09s & 2.0B & 0.84 & 0.39 & 0.60 & 0.72 & 84.19 & 93.07 & 82.94 \\
        \rowcolor{lightblue}
        +NOVA & 2.89 $\times$ & 0.84s & 2.0B & 0.86 & 0.42 & 0.58 & 0.72 & 84.36 & 92.68 & 82.93 \\
        \midrule
        Infinity-8B & 1.00 $\times$ & 1.51s & 8.0B & 0.94 & 0.58 & 0.68 & 0.80 & 85.10 & 94.50 & 86.60 \\
        +FastVAR & 1.26 $\times$ & 0.79s & 8.0B & 0.92 & 0.55 & 0.68 & 0.78 & 84.80 & 92.30 & 86.46 \\
        +SkipVAR & 1.22 $\times$ & 0.82s & 8.0B & 0.91 & 0.57 & 0.69 & 0.77 & 83.91 & 90.52 & 86.40 \\
        \rowcolor{lightblue}
        +NOVA & 1.33 $\times$ & 0.75s & 8.0B & 0.92 & 0.58 & 0.69 & 0.79 & 84.27 & 93.61 & 86.53 \\
        
        \bottomrule
    \end{tabular}
    \vspace{-0.4cm}
\end{table*}

For layer-linkage ratio function, we dynamically refine the scale reduction ratio by computing the relative deviation between the current layer’s overall entropy and the mean overall entropy of all preceding layers and use this as the criterion for adjusting the token reduction ratio at the current layer. For the $t$-th scale and $j$-th layer, the mean overall entropy $\mu_{t,j-1}$ of the preceding $j-1$ layers is:
\begin{equation}
\mu_{t,j-1}=
\frac{1}{j-1}\sum_{k=1}^{j-1}\bar{\mathcal{H}}_{t,k},
\quad j\ge 2.
\label{eq:prev_layer_mean}
\end{equation}
Therefore, $\Delta_{t,j}=(\bar{\mathcal{H}}_{t,j}-\mu_{t,j-1})/\mu_{t,j-1}$ is the adjustment  for the $j$-th layer's overall entropy $\bar{\mathcal{H}}_{t,j}$. The token reduction ratio $\mathrm{Ratio}_{t,j}$ at the $t$-th scale and $j$-th layer is:
\begin{equation}
\mathrm{Ratio}_{t,j}
=
\begin{cases}
\mathrm{Ratio}_{t}, & j = 1,\\[4pt]
\mathrm{Ratio}_{t}-\Delta_{t,j}\mathrm{Ratio}_{t} , & j \ge 2.
\end{cases}
\label{eq:layer_linkage_piecewise}
\end{equation}
Finally, pruning low-entropy tokens according to the token reduction ratio can achieve dual-linkage acceleration.

\paragraph{Residual Cache Reuse.}
As a dense prediction task, visual generation requires preserving a coherent two-dimensional structure, we employ residual cache reuse to compensate for pruned low-entropy tokens. At scales with $t \geq t^\ast$, the residual cache is activated and updated. The layer transformation at the $j$-th layer (Transformer layer $j$) for the $t$-th scale can be expressed as:
\begin{equation}
    R^{\text{output}}_{t,j} = \mathrm{Layer}_j\!\left(R^{\text{input}}_{t,j}\right),
\end{equation}
Consequently, we construct a layer-wise residual cache:
\begin{equation}
    \mathrm{Cache}_{t,j} = R^{\text{output}}_{t,j} - R^{\text{input}}_{t,j}.
\end{equation}
When predicting the next scale $t+1$ at $j$-th layer, for those pruned tokens, we interpolate $\mathrm{Cache}_{t,j}$ into $\mathrm{Cache}_{t\rightarrow t+1,j}$ and reconstruct the layer output by reusing residual cache:
\begin{equation}
R^{\text{output}}_{t,j}=\mathrm{Cache}_{t\rightarrow t+1,j}+R^{\text{input}}_{t,j}.
\end{equation}
By reusing dynamic residual information rather than static tokens, NOVA preserves fine-grained structural cues while avoiding redundant computation on low-entropy tokens.

\section{Experiments}

\subsection{Experimental Settings}

\paragraph{Models and Benchmarks.}

We employ text-to-image VAR models to evaluate NOVA, including Infinity-2B, Infinity-8B~\cite{han2025infinity}, and HART-0.7B~\cite{tang2024hart}. All selected backbones are capable of high-resolution image generation at $1024\times1024$. We utilize GenEval~\cite{ghosh2023geneval}, DPG-Bench~\cite{dpgbench}, and MJHQ-30K~\cite{li2024playground} to evaluate generation quality, which are widely adopted for assessing semantic alignment and the perceptual quality of generated images. Moreover, we use ImageReward~\cite{xu2023imagereward} and HPSv2.1~\cite{wu2023human}. 
To quantify efficiency, we report the average inference latency and the corresponding speedup ratio achieved by our method.

\begin{table*}[t]
    \centering
    \caption{Quantitative evaluation on ImageReward and HPSv2.1, two widely used human preference benchmarks. }
    \vspace{-0.2cm}
    \label{tab:evaluation:human}
    \resizebox{\textwidth}{!}{
    \begin{tabular}{lcccccccc}
        \toprule
        \multirow{2}{*}{\textbf{Methods}} & \multirow{2}{*}{\textbf{Latency}} & \multicolumn{2}{c}{\textbf{ImageReward}} & \multicolumn{5}{c}{\textbf{HPSv2.1}} \\
        \cmidrule(lr){3-4} \cmidrule(lr){5-9}
         & & \textbf{Score}$\uparrow$  & \textbf{CLIP-Score}$\uparrow$ & \textbf{Anime}$\uparrow$ & \textbf{Concept-Art}$\uparrow$ & \textbf{Paintings}$\uparrow$ & \textbf{Photo}$\uparrow$ & \textbf{Overall}$\uparrow$ \\
        \midrule
        Infinity-2B & 2.43s & 0.921 & 0.27 & 31.63 & 30.26 & 30.28 & 29.27 & 30.36 \\
        +FastVAR & 0.99s & 0.903 & 0.26 & 31.12 & 30.15 & 29.92 & 28.86 & 29.95 \\
        +SkipVAR & 1.09s & 0.908 & 0.27 & 30.11 & 30.18 & 30.13 & 29.07 & 30.22  \\
        +SparseVAR & 1.46s & 0.895 & 0.26 & 31.22 & 29.61 & 29.10 & 28.21 & 29.53 \\
        \rowcolor{lightblue}
        +NOVA & 0.84s & 0.911 & 0.27 & 31.40 & 30.06 & 30.21 & 29.11 & 30.20 \\
        \midrule
        Infinity-8B  & 1.51s &  1.030 & 0.28 & 32.40   &   31.17   &   30.78   &   29.35   &   30.92   \\
        +FastVAR &  0.79s  & 1.028 &  0.28   &   31.80   &   30.42   &   29.89   &   29.87   &   30.24   \\
        +SkipVAR  &  0.82s   & 1.032 &  0.27     &   32.01   &   30.58   &   30.46   &   29.09   &   30.44   \\
        \rowcolor{lightblue}
        +NOVA   & 0.75s &  1.035    &  0.28   &   32.03   &   30.71   &   30.25   &   29.15   &   30.54   \\
        \bottomrule
    \end{tabular}
    }
    \vspace{-0.3cm}
\end{table*}

\paragraph{Implementation Details.}

To ensure a fair comparison, we evaluate NOVA against all open-source baselines using their default hyperparameter settings. For NOVA, we set the adaptive acceleration activation parameters to $t_{\mathrm{est}}=5$ and $\alpha=0.5$, and set the dual-linkage parameters to $\tau=0.8$ and $\lambda=0.1$. These hyperparameters are not heavily tuned for each benchmark; we use nearly identical settings across different backbones. The sensitivity analysis in appendix Table~\ref{tab:ablation_tau} and appendix Figure~\ref{fig:lambda_ablation} show stable GenEval performance for $\tau\in[0.6,0.9]$ and $\lambda\in[0.05,0.25]$, indicating limited tuning overhead. More implementation details are provided in the appendix. All experiments are conducted on a single NVIDIA RTX 3090 GPU, except for Infinity-8B, which is evaluated on an NVIDIA H200 GPU due to its larger model size.

\subsection{Main Results}
\paragraph{Quantitative Performance Comparison.}
As shown in Table~\ref{tab:comparison} and Table~\ref{tab:main:mjhq}, we assess both generation quality and inference efficiency for $1024 \times 1024$ image generation on GenEval, DPG-Bench, and MJHQ-30K. Compared with other comparable models, NOVA consistently achieves a superior balance of efficiency and performance. Specifically, for Infinity-2B, NOVA delivers $2.89\times$ speedup while preserving approximately original GenEval Overall score 0.72. Notably, NOVA attains the best performance on Two Object sub-metric, outperforming both the original Infinity-2B and other acceleration methods. Moreover, on DPG-Bench and MJHQ-30K, results also confirm that NOVA preserves high-fidelity visual quality under acceleration.  Furthermore, on larger-size and other VAR models such as Infinity-8B and HART, NOVA consistently delivers substantial acceleration, demonstrating its broad applicability and generalizability across diverse model architectures and sizes.

\begin{table}[h]
    \centering
    \caption{Performance on GenEval$\uparrow$ benchmark.}
    \label{tab:main:mjhq}
    \resizebox{0.48\textwidth}{!}{
    \begin{tabular}{l|cc|cc|cc}
        \toprule
        \multirow{2}{*}{\textbf{Methods}} 
        & \multicolumn{2}{c}{\textbf{Landscape}} 
        & \multicolumn{2}{c}{\textbf{People}} 
        & \multicolumn{2}{c}{\textbf{Food}} \\
        \cmidrule(lr){2-3} \cmidrule(lr){4-5} \cmidrule(lr){6-7}
        & \textbf{FID}$\downarrow$ & \textbf{CLIP}$\uparrow$
        & \textbf{FID}$\downarrow$ & \textbf{CLIP}$\uparrow$
        & \textbf{FID}$\downarrow$ & \textbf{CLIP}$\uparrow$ \\
        \midrule
        HART        & 25.43 & 26.82 & 30.61 & 28.47 & 30.37 & 28.03 \\
        +FastVAR     & 22.52 & 26.51 & 28.19 & 28.34 & 30.97 & 28.25 \\
        \rowcolor{lightblue}
        +NOVA        & 22.81 & 26.56 & 28.09 & 28.33 & 30.74 & 28.40 \\
        \midrule
        Infinity-2B & 29.29 & 26.18 & 28.15 & 27.97 & 30.73 & 26.70 \\
        +FastVAR     & 28.08 & 26.58 & 27.99 & 28.31 & 29.95 & 27.08 \\
        \rowcolor{lightblue}
        +NOVA        & 29.29 & 26.14 & 27.88 & 29.29 & 29.91 & 27.03 \\
        \bottomrule
    \end{tabular}
    }
\end{table}

\paragraph{Quantitative Performance on Human Preference.}
We further evaluate the impact of NOVA's acceleration on perceptual quality using human preference benchmarks with ImageReward and HPSv2.1. In Table~\ref{tab:evaluation:human}, despite the aggressive reduction in token usage, NOVA maintains high alignment with human aesthetic preferences. Surprisingly, on ImageReward, NOVA obtains a score of 1.035 on Infinity-8B, outperforming the initial Infinity-8B (1.030). This breaks the conventional wisdom that accelerating through token pruning necessarily leads to performance loss. On HPSv2.1, NOVA remains comparable to the unpruned baseline, confirming that the generated images are free of perceptible artifacts and maintain high artistic quality.

\begin{figure}[t]
    \centering
    \includegraphics[width=1\linewidth]{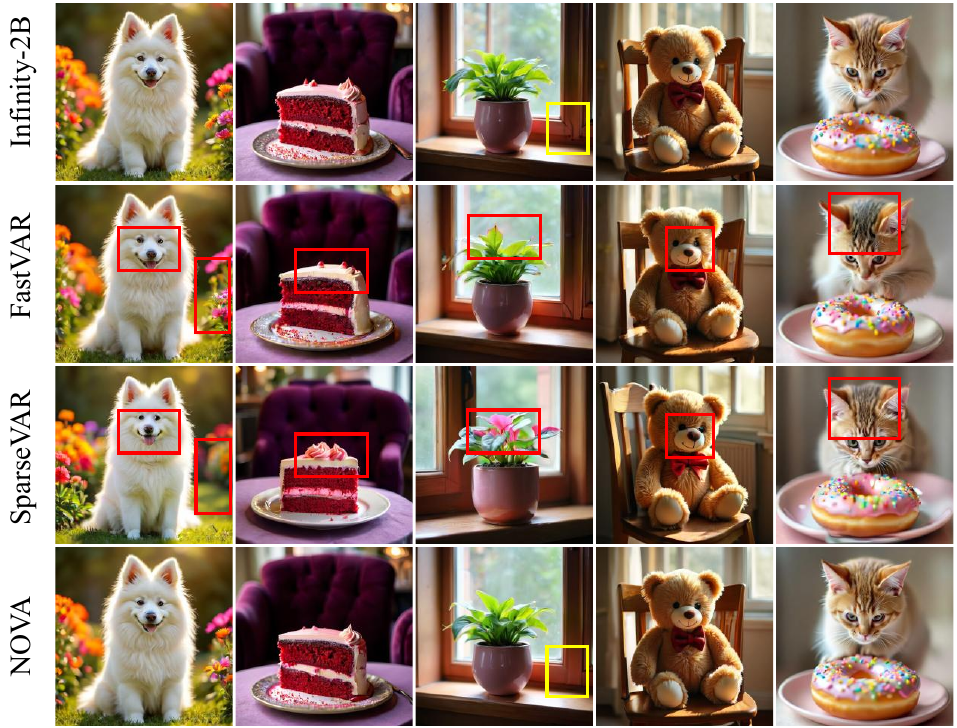}
    \caption{Visual comparison between NOVA and other training-free comparable methods on Infinity-2B. \textcolor{red}{Red} and \textcolor{yellow}{yellow} boxes highlight the finer details in visualization.
}
    \label{fig:infinity_2bcompare}
    \vspace{-0.5cm}
\end{figure}

\begin{figure*}[t]
    \centering
    \includegraphics[width=1\textwidth]{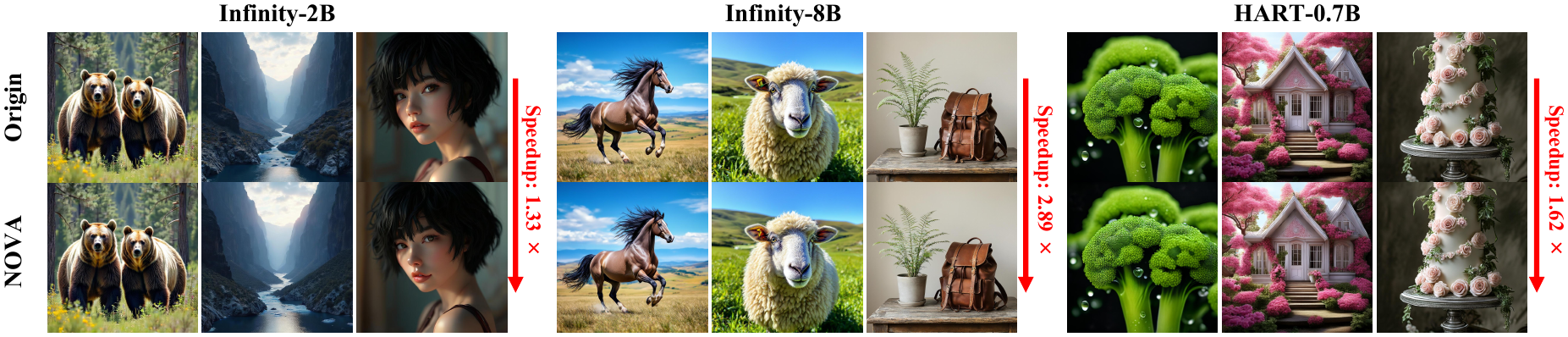}
    \vspace{-0.5cm}
    \caption{Visual comparisons between NOVA and backbones, including Infinity-2B, Infinity-8B, and HART-0.7B.}
    \label{fig:various_compare}
    \vspace{-0.4cm}
\end{figure*}

\paragraph{Visual Comparison.}
In Figure~\ref{fig:various_compare} and Figure~\ref{fig:infinity_2bcompare}, we provide visual comparisons to illustrate the performance of NOVA across different backbones and against comparable token reduction acceleration methods. These results indicate that under the adaptive acceleration activation and dual-linkage acceleration, NOVA achieves near-lossless visual quality mitigates semantic loss, structural distortion, and detail collapse. Notably, in the yellow-box region, NOVA generates clearer and more logically consistent structural details than the original Infinity-2B, aligning better with human preference. More comparisons are provided in the appendix.

\begin{table}[h]
\vspace{-0.2cm}
    \centering
    
    \caption{Ablation of token selection metrics on Infinity-2B.}
    \vspace{-0.15cm}
    \label{tab:aba:entropy}
    \resizebox{0.48\textwidth}{!}{
        \begin{tabular}{l|cccc} 
            \toprule
            \multirow{2}{*}{\textbf{Methods}} & \multicolumn{4}{c}{\textbf{GenEval} $\uparrow$} \\
            \cmidrule(lr){2-5}
            & Two Obj. & Position & Color Attri. & \textbf{Overall} \\
            \midrule
            Infinity-2B & 0.85 & 0.45 & 0.54 & 0.73 \\
            +NOVA w/ Attn & 0.84 & 0.38 & 0.52 & 0.69 \\
            +NOVA w/ MSE & 0.82 & 0.42 & 0.52 & 0.71 \\
            \rowcolor{lightblue}
            +NOVA w/ Entropy & 0.86 & 0.42 & 0.54 & 0.72 \\
            \bottomrule
        \end{tabular}
    }
    \vspace{-0.3cm}
\end{table}

\subsection{Ablation and Analysis}
\paragraph{Effectiveness of Entropy Analysis.}
To examine the effectiveness of our entropy-based method for token reduction, we conduct an ablation study comparing it with Attention-based (Attn) and MSE-based (MSE) alternatives on Infinity-2B. All variants adopt the same dual-linkage configuration for fair comparison. As shown in Table~\ref{tab:aba:entropy}, the MSE-based method leads to a noticeable drop in overall GenEval performance, while the entropy-based variant achieves the best overall score among all compared methods. Similar to SparseVAR~\cite{chen2025frequency}, we treat MSE metric as a proxy for frequency signal magnitude. These results suggest that frequency signals alone may not reliably reflect their impact on potential information, whereas entropy provides a more informative criterion for identifying tokens that contribute important information in the pruning process.
\begin{table}[h]
\vspace{-0.15cm}
    \centering
    \caption{Ablation on the effectiveness of adaptive acceleration activation strategy on Infinity-2B. The `*' denote without adaptive acceleration activation.}
    \label{tab:aba:aaa}
    \resizebox{0.48\textwidth}{!}{
        \begin{tabular}{l|c|c|ccc}
            \toprule
            \multirow{2}{*}{\textbf{Methods}} 
            & \multirow{2}{*}{\textbf{Start Scale}} 
            & \multirow{2}{*}{\textbf{Speedup}} 
            & \multicolumn{3}{c}{\textbf{GenEval} $\uparrow$} \\
            \cmidrule(lr){4-6}
            & & & Two Obj. & Color Attri. & \textbf{Overall} \\
            \midrule
            NOVA* & 6-th & 3.52 $\times$ & 0.72 & 0.45 & 0.62 \\
            NOVA* & 7-th & 2.85 $\times$ & 0.78 & 0.49 & 0.66 \\
            NOVA* & 8-th & 2.14 $\times$ & 0.81 & 0.53 & 0.68 \\
            NOVA*& 9-th & 1.88 $\times$ & 0.84 & 0.54 & 0.70 \\
            NOVA* & 10-th & 1.63 $\times$ & 0.83 & 0.54 & 0.70 \\
            \rowcolor{lightblue}
            NOVA & -- & 2.89 $\times$ & 0.86 & 0.58 & 0.72 \\
            \bottomrule
        \end{tabular}
    }
    \vspace{-0.1cm}
\end{table}

\paragraph{Effectiveness of Adaptive Acceleration Activation.}
To demonstrate the effectiveness of the adaptive acceleration activation, we compare NOVA with fixed acceleration scales on Infinity-2B. To ensure a fair comparison, all methods adopt the adaptive dual-linkage ratio function to determine reduction ratio at each scale. As shown in Table~\ref{tab:aba:aaa}, set the fixed acceleration scale early (e.g., 6-th scale) may lead to aggressive speedup but causes noticeable quality degradation, whereas set a delaying acceleration scale (e.g., 10-th scale) preserves quality at the cost of reduced acceleration. In contrast, NOVA adaptively determines the acceleration scale, enabling to achieve a more favorable trade-off between aggressive acceleration and quality preservation.

\begin{table}[h]
    \centering
    \caption{Ablation on effectiveness of dual-linkage acceleration on Infinity-2B. $\dagger$ denotes the variants' scale-level reduction ratios are fixed following FastVAR, $\ddagger$ denotes the variants' scale-level reduction ratios following SparseVAR.}
    \label{tab:aba:dual_linkage}
    \resizebox{0.48\textwidth}{!}{
        \begin{tabular}{l|c|c|cc}
            \toprule
            \textbf{Methods} & \textbf{Scale-Linkage} & \textbf{Layer-Linkage} & \textbf{Speedup} & \textbf{GenEval} $\uparrow$ \\
            \midrule
            NOVA$^{\dagger}$ & \redcross & \redcross & 2.55 $\times$ & 0.69 \\
            NOVA$^{\dagger}$ & \redcross & \greencheck & 2.34 $\times$ & 0.70 \\
            NOVA$^{\ddagger}$ & \redcross & \redcross & 2.15 $\times$ & 0.71 \\
            NOVA$^{\ddagger}$ & \redcross & \greencheck & 2.22 $\times$ & 0.71 \\
            NOVA & \greencheck & \redcross & 2.85 $\times$ & 0.71 \\
            \rowcolor{lightblue}
            NOVA& \greencheck & \greencheck & 2.89 $\times$ & 0.72 \\
            \bottomrule
        \end{tabular}
    }
    \vspace{-0.4cm}
\end{table}

\paragraph{Effectiveness of Dual-Linkage Acceleration.}

We conduct an ablation study covering two no-linkage and three single-linkage variants to examine the effectiveness of dual-linkage acceleration. All variants employ adaptive acceleration activation to ensure a fair comparison. As shown in Table~\ref{tab:aba:dual_linkage}, both no-linkage variants suffer from an inherent efficiency-quality trade-off. Further, enabling either scale-linkage or layer-linkage alone consistently improves over the corresponding no-linkage variants, while jointly applying dual-linkage pushes this improvement to the best overall efficiency and performance trade-off.

\begin{table}[h]
    \centering
    \caption{Ablation on the effectiveness of residual cache reuse strategy on Infinity-2B. The `*' denotes static token caching strategy.}
    \label{tab:rcr}
    \resizebox{0.48\textwidth}{!}{
        \begin{tabular}{l|c|c|ccc}
            \toprule
            \multirow{2}{*}{\textbf{Methods}} 
            & \multirow{2}{*}{\textbf{Start Cache Scale}} 
            & \multirow{2}{*}{\textbf{Speedup}} 
            & \multicolumn{3}{c}{\textbf{GenEval} $\uparrow$} \\
            \cmidrule(lr){4-6}
            & & & Two Obj. & Color Attri. & \textbf{Overall} \\
            \midrule
            NOVA* & 6-th & 2.93 $\times$ & 0.78 & 0.49 & 0.67 \\
            NOVA* & 7-th & 2.77 $\times$ & 0.79 & 0.45 & 0.69 \\
            NOVA* & 8-th & 2.33 $\times$ & 0.81 & 0.53 & 0.71 \\
            \rowcolor{lightblue}
            NOVA & -- & 2.89 $\times$ & 0.86 & 0.58 & 0.72 \\
            \bottomrule
        \end{tabular}
    }
    \vspace{-0.2cm}
\end{table}

\paragraph{Effectiveness of Residual Cache Reuse.}
To verify the effectiveness of the proposed residual cache reuse strategy, we conduct an ablation study on the Infinity-2B backbone. We compare NOVA's dynamic residual cache reuse with FastVAR's static token caching, which caches tokens at a predetermined decoding step. For a fair comparison across cache scales, $t_{\text{est}}$ is adjusted accordingly for both methods under comparable acceleration settings. Table~\ref{tab:rcr} shows that dynamic residual cache reuse consistently yields better generation quality under similar or even higher speedup ratios. Unlike static caching, which reuses tokens at fixed decoding steps regardless of the modeling state, NOVA adaptively reuses residual features via entropy-guided token selection. This dynamic behavior better preserves semantic consistency and fine-grained details, while mitigating error accumulation from aggressive token pruning.

\begin{table}[h]
    \vspace{-0.2cm}
    \centering
    \caption{Latency breakdown of a representative high-resolution scale during VAR-like model inference on the Infinity-2B backbone at $1024\times1024$ resolution.}
    \label{tab:entropy_overhead}
    \renewcommand{\arraystretch}{0.85}
    \resizebox{0.48\textwidth}{!}{
        \begin{tabular}{lcc}
            \toprule
            \textbf{Component} & \textbf{Latency (ms)} & \textbf{Ratio (\%)} \\
            \midrule
            Self-Attention      & 128.4 & 71.2 \\
            Feed-Forward (FFN)  & 42.1  & 23.4 \\
            Entropy Computation & 1.7   & 0.9  \\
            Others              & 8.2   & 4.5  \\
            \midrule
            Total               & 180.4 & 100.0 \\
            \bottomrule
        \end{tabular}
    }
    \vspace{-0.4cm}
\end{table}

\paragraph{Entropy Computation Overhead.}
We further analyze the runtime overhead introduced by entropy computation. NOVA computes entropy directly from the predictive token distributions already produced during standard decoding, requiring only element-wise operations and reductions. The additional cost scales as $\mathcal{O}(N_t |V|)$ at scale $t$, which is linear in the number of tokens, while self-attention scales as $\mathcal{O}(N_t^2 d)$. In practice, entropy computation takes only 1.7 ms on a representative high-resolution scale of Infinity-2B at $1024 \times 1024$ resolution, accounting for 0.9\% of the total latency; the full latency breakdown is provided in Appendix~\ref{entropy_app}. Therefore, the entropy overhead is small and is effectively amortized by the computation saved through token reduction.

\begin{table}[h]
    \centering
    \caption{Quantitative comparison on class-conditional ImageNet generation using VAR-d30 at $256\times256$ and VAR-d36 at $512\times512$ resolution. Latency is measured with batch size 5. FID and IS are reported to evaluate generation quality.}
    \label{tab:c2i}
    \resizebox{0.48\textwidth}{!}{
    \begin{tabular}{l|cccc}
        \toprule
        \textbf{Method} & \textbf{Res.} & \textbf{Latency} $\downarrow$ & \textbf{FID} $\downarrow$  & \textbf{IS} $\uparrow$\\
        \midrule
        VAR-d30 & 256 & 0.67s & 2.05 & 306.6 \\
        VAR-d36 & 512 & 2.89s & 2.63 & 303.2 \\
        \rowcolor{lightblue}
        NOVA(VAR-d30) & 256 & 0.63s & 2.23 & 313.2 \\
        \rowcolor{lightblue}
        NOVA(VAR-d36) & 512 & 1.97s & 2.79 & 299.1 \\
        \bottomrule
    \end{tabular}
    }
    \vspace{-0.3cm}
\end{table}

\paragraph{Class-Conditional Generation.}
To examine the generality of NOVA beyond text-to-image generation, we further evaluate it on class-conditional ImageNet generation using VAR-d30 at $256\times256$ resolution and VAR-d36 at $512\times512$ resolution, with the batch size fixed to 5. As shown in Table~\ref{tab:c2i}, NOVA reduces the latency of VAR-d36 from 2.89s to 1.97s with only a modest change in FID and IS. The acceleration on VAR-d30 is less pronounced, reducing latency from 0.67s to 0.63s, likely because the lower-resolution model has less computational redundancy to exploit. These results suggest that NOVA remains effective in class-conditional generation, especially for higher-resolution settings with larger computational cost.

\begin{figure}[h]
    \centering
    \includegraphics[width=1\linewidth]{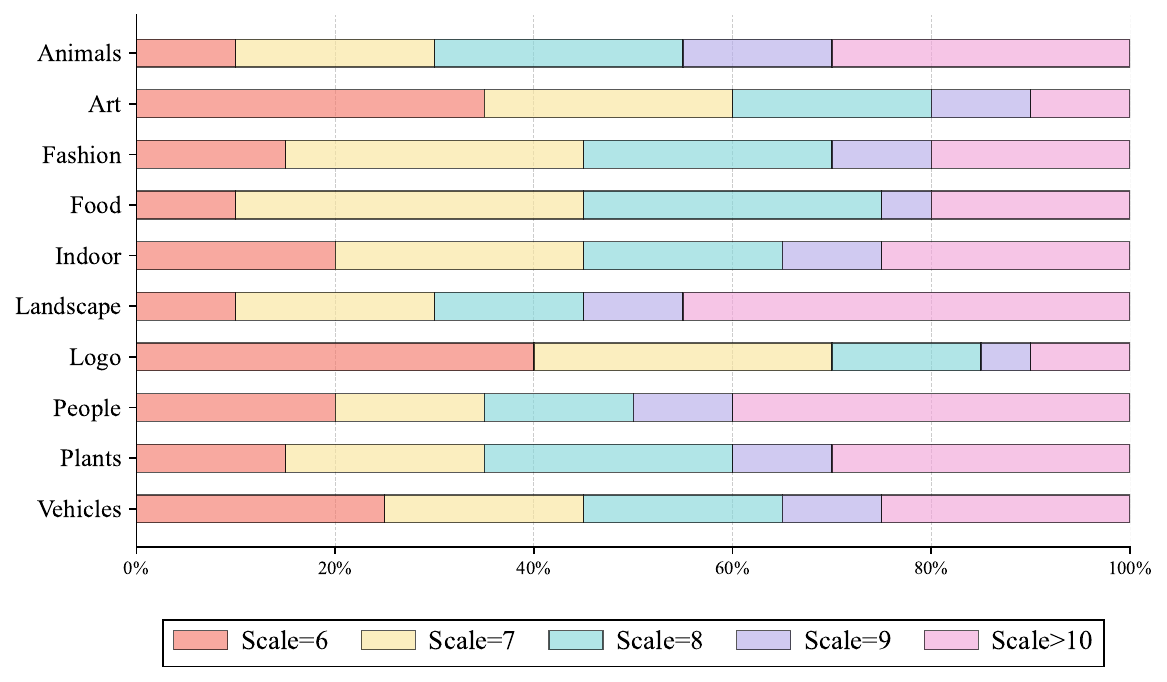}
    \vspace{-0.6cm}
    \caption{Distribution of acceleration activation scale ratios of NOVA across different categories of the MJHQ-30K benchmark.}
    \label{fig:statistic}
    \vspace{-0.5cm}
\end{figure}

\paragraph{Statistical Analysis of Acceleration Activation Scale.}
We statistically analyze the acceleration activation scale on MJHQ-30K by examining its distribution across image categories.
In Figure~\ref{fig:statistic}, diverse image categories exhibit distinct activation distributions, indicating that pruning onset depends on content: fine-grained textures necessitate later pruning, whereas simple global structures allow earlier activation. The results suggest that NOVA can effectively capture context-specific generation characteristics and adaptively identify appropriate acceleration scales, rather than relying on fixed or heuristic reduction schedules. This instance-level variation further supports the need for adaptive acceleration rather than a fixed pruning schedule in VAR inference.

\section{Conclusion}
This paper introduces NOVA, a training-free acceleration framework for VAR models. By utilizing dual-linkage entropy analysis, NOVA adaptively online identifies the acceleration activation scale through entropy growth inflection point detection. NOVA also dynamically computes distinct token reduction ratios across both scales and layers, pruning low-entropy tokens and reusing residual cache to accelerate inference and maintain quality. Extensive experiments validate NOVA as a simple yet effective acceleration method. 

\section*{Acknowledgement}
This work is supported by the National Natural Science Foundation of China (No. 62576251, No. 62406225, No. 62376198, and No. 62576247), the National Science and Technology Major Project of China (No. 2025ZD0219200), and the Shanghai Science and Technology Committee under Grant No. 24511103900.

\clearpage
\newpage

\section*{Impact Statement}
This paper presents work whose goal is to advance the field of Machine
Learning. There are many potential societal consequences of our work, none which we feel must be specifically highlighted here.

\bibliography{example_paper}

@inproceedings{koumll,
	title     = {Exploiting Multi-Label Correlation in Label Distribution Learning},
	author    = {Kou, Zhiqiang and Wang, Jing and Tang, Jiawei and Jia, Yuheng and Shi, Boyu and Geng, Xin},
	booktitle = {Proceedings of the Thirty-Third International Joint Conference on Artificial Intelligence},

	pages     = {4326--4334},
	year      = {2024},
	month     = {8}
}

@inproceedings{koubldl,
	title     = {Label Distribution Learning with Biased Annotations Assisted by Multi-Label Learning},
	author    = {Kou, Zhiqiang and Qin, Si and Wang, Hailin and Wang, Jing and Xie, Mingkun and Chen, Shuo and Jia, Yuheng and Liu, Tongliang and Sugiyama, Masashi and Geng, Xin},
	booktitle = {Proceedings of the Thirty-Fourth International Joint Conference on Artificial Intelligence},
	year      = {2025},
	month     = {8}
}

@inproceedings{kou2025nips,
  title     = {RankMatch: A Novel Approach to Semi-Supervised Label Distribution Learning Leveraging Rank Correlation between Labels},
  author    = {Zhiqiang Kou and Yucheng Xie and Hailin Wang and Jing Wang and Mingkun Xie and Shuo Chen and Yuheng Jia and Tongliang Liu and Xin Geng},
  booktitle = {Proceedings of the 39th Conference on Neural Information Processing Systems (NeurIPS 2025)},
  year      = {2025},
  address   = {San Diego, California, USA},
  month     = {December},
  organization = {Neural Information Processing Systems Foundation},
 
}

@inproceedings{kou2026fedharmony,
  title     = {FedHarmony: Harmonizing Heterogeneous Label Correlations in Federated Multi-Label Learning},
  author    = {Kou, Zhiqiang and Wu, Junxiang and Huang, Wenke and He, Wenwen and Xie, Ming-Kun and Wang, Changwei and Jia, Yuheng and Jiang, Di and Liu, Yang and Geng, Xin and Yang, Qiang},
  booktitle = {Proceedings of the IEEE/CVF Conference on Computer Vision and Pattern Recognition (CVPR)},
  year      = {2026}
}

@misc{kou2026pu,
      title={Positive-Unlabeled Reinforcement Learning Distillation for On-Premise Small Models}, 
      author={Zhiqiang Kou and Junyang Chen and Xin-Qiang Cai and Xiaobo Xia and Ming-Kun Xie and Dong-Dong Wu and Biao Liu and Yuheng Jia and Xin Geng and Masashi Sugiyama and Tat-Seng Chua},
      year={2026},
      eprint={2601.20687},
      archivePrefix={arXiv},
      primaryClass={cs.LG},
      url={https://arxiv.org/abs/2601.20687}, 
}

@misc{wu2026trustworthyfederatedlabeldistribution,
      title={Trustworthy Federated Label Distribution Learning under Annotation Quality Disparity}, 
      author={Junxiang Wu and Zhiqiang Kou and Hongwei Zeng and Wenke Huang and Biao Liu and Hanlin Gu and Yuheng Jia and Di Jiang and Yang Liu and Xin Geng},
      year={2026},
      eprint={2605.04827},
      archivePrefix={arXiv},
      primaryClass={cs.LG},
      url={https://arxiv.org/abs/2605.04827}, 
}

@ARTICLE{koutnnls,
  author={Kou, Zhiqiang and Wang, Jing and Jia, Yuheng and Liu, Biao and Geng, Xin},
  journal={IEEE Transactions on Neural Networks and Learning Systems}, 
  title={Instance-Dependent Inaccurate Label Distribution Learning}, 
  year={2025},
  volume={36},
  number={1},
  pages={1425-1437},
  doi={10.1109/TNNLS.2023.3329870}}

@article{tian2024visual,
  title={Visual autoregressive modeling: Scalable image generation via next-scale prediction},
  author={Tian, Keyu and Jiang, Yi and Yuan, Zehuan and Peng, Bingyue and Wang, Liwei},
  journal={Advances in neural information processing systems},
  volume={37},
  pages={84839--84865},
  year={2024}
}

@article{croitoru2023diffusion,
  title={Diffusion models in vision: A survey},
  author={Croitoru, Florinel-Alin and Hondru, Vlad and Ionescu, Radu Tudor and Shah, Mubarak},
  journal={IEEE transactions on pattern analysis and machine intelligence},
  volume={45},
  number={9},
  pages={10850--10869},
  year={2023},
  publisher={Ieee}
}

@article{shen2025efficient,
  title={Efficient diffusion models: A survey},
  author={Shen, Hui and Zhang, Jingxuan and Xiong, Boning and Hu, Rui and Chen, Shoufa and Wan, Zhongwei and Wang, Xin and Zhang, Yu and Gong, Zixuan and Bao, Guangyin and others},
  journal={arXiv preprint arXiv:2502.06805},
  year={2025}
}

@article{cao2024survey,
  title={A survey on generative diffusion models},
  author={Cao, Hanqun and Tan, Cheng and Gao, Zhangyang and Xu, Yilun and Chen, Guangyong and Heng, Pheng-Ann and Li, Stan Z},
  journal={IEEE transactions on knowledge and data engineering},
  volume={36},
  number={7},
  pages={2814--2830},
  year={2024},
  publisher={IEEE}
}

@article{ren2024m,
  title={M-var: Decoupled scale-wise autoregressive modeling for high-quality image generation},
  author={Ren, Sucheng and Yu, Yaodong and Ruiz, Nataniel and Wang, Feng and Yuille, Alan and Xie, Cihang},
  journal={arXiv preprint arXiv:2411.10433},
  year={2024}
}

@inproceedings{kumbong2025hmar,
  title={HMAR: Efficient Hierarchical Masked Auto-Regressive Image Generation},
  author={Kumbong, Hermann and Liu, Xian and Lin, Tsung-Yi and Liu, Ming-Yu and Liu, Xihui and Liu, Ziwei and Fu, Daniel Y and Re, Christopher and Romero, David W},
  booktitle={Proceedings of the Computer Vision and Pattern Recognition Conference},
  pages={2535--2544},
  year={2025}
}

@article{huang2025spectralar,
  title={SpectralAR: Spectral Autoregressive Visual Generation},
  author={Huang, Yuanhui and Chen, Weiliang and Zheng, Wenzhao and Duan, Yueqi and Zhou, Jie and Lu, Jiwen},
  journal={arXiv preprint arXiv:2506.10962},
  year={2025}
}

@article{he2025neighboring,
  title={Neighboring autoregressive modeling for efficient visual generation},
  author={He, Yefei and He, Yuanyu and He, Shaoxuan and Chen, Feng and Zhou, Hong and Zhang, Kaipeng and Zhuang, Bohan},
  journal={arXiv preprint arXiv:2503.10696},
  year={2025}
}

@inproceedings{chen2025collaborative,
  title={Collaborative decoding makes visual auto-regressive modeling efficient},
  author={Chen, Zigeng and Ma, Xinyin and Fang, Gongfan and Wang, Xinchao},
  booktitle={Proceedings of the Computer Vision and Pattern Recognition Conference},
  pages={23334--23344},
  year={2025}
}

@article{kumar2025scale,
  title={Scale-wise var is secretly discrete diffusion},
  author={Kumar, Amandeep and Nair, Nithin Gopalakrishnan and Patel, Vishal M},
  journal={arXiv preprint arXiv:2509.22636},
  year={2025}
}

@article{chen2025tts,
  title={Tts-var: A test-time scaling framework for visual auto-regressive generation},
  author={Chen, Zhekai and Chu, Ruihang and Chen, Yukang and Zhang, Shiwei and Wei, Yujie and Zhang, Yingya and Liu, Xihui},
  journal={arXiv preprint arXiv:2507.18537},
  year={2025}
}

@article{vincenti2025dynamic,
  title={Dynamic Mixture-of-Experts for Visual Autoregressive Model},
  author={Vincenti, Jort and Jazbec, Metod and Xia, Guoxuan},
  journal={arXiv preprint arXiv:2510.08629},
  year={2025}
}

@article{tang2024hart,
  title={Hart: Efficient visual generation with hybrid autoregressive transformer},
  author={Tang, Haotian and Wu, Yecheng and Yang, Shang and Xie, Enze and Chen, Junsong and Chen, Junyu and Zhang, Zhuoyang and Cai, Han and Lu, Yao and Han, Song},
  journal={arXiv preprint arXiv:2410.10812},
  year={2024}
}

@article{ma2024star,
  title={STAR: Scale-wise Text-conditioned AutoRegressive image generation},
  author={Ma, Xiaoxiao and Zhou, Mohan and Liang, Tao and Bai, Yalong and Zhao, Tiejun and Li, Biye and Chen, Huaian and Jin, Yi},
  journal={arXiv preprint arXiv:2406.10797},
  year={2024}
}

@inproceedings{wu2023sega,
  title={Sega: Structural entropy guided anchor view for graph contrastive learning},
  author={Wu, Junran and Chen, Xueyuan and Shi, Bowen and Li, Shangzhe and Xu, Ke},
  booktitle={International Conference on Machine Learning},
  pages={37293--37312},
  year={2023},
  organization={PMLR}
}

@inproceedings{yang2023minimum,
  title={Minimum entropy principle guided graph neural networks},
  author={Yang, Zhenyu and Zhang, Ge and Wu, Jia and Yang, Jian and Sheng, Quan Z and Peng, Hao and Li, Angsheng and Xue, Shan and Su, Jianlin},
  booktitle={Proceedings of the sixteenth ACM international conference on web search and data mining},
  pages={114--122},
  year={2023}
}

@inproceedings{yang2024sebot,
  title={Sebot: Structural entropy guided multi-view contrastive learning for social bot detection},
  author={Yang, Yingguang and Wu, Qi and He, Buyun and Peng, Hao and Yang, Renyu and Hao, Zhifeng and Liao, Yong},
  booktitle={Proceedings of the 30th ACM SIGKDD conference on knowledge discovery and data mining},
  pages={3841--3852},
  year={2024}
}

@article{wang2023edrl,
  title={EDRL: Entropy-guided disentangled representation learning for unsupervised domain adaptation in semantic segmentation},
  author={Wang, Runze and Zhou, Qin and Zheng, Guoyan},
  journal={Computer methods and programs in biomedicine},
  volume={240},
  pages={107729},
  year={2023},
  publisher={Elsevier}
}

@inproceedings{xiao2024eggesture,
  title={Eggesture: Entropy-guided vector quantized variational autoencoder for co-speech gesture generation},
  author={Xiao, Yiyong and Shu, Kai and Zhang, Haoyi and Yin, Baohua and Cheang, Wai Seng and Wang, Haoyang and Gao, Jiechao},
  booktitle={Proceedings of the 32nd ACM International Conference on Multimedia},
  pages={6113--6122},
  year={2024}
}

@article{li2025efficient,
  title={Efficient Training for Human Video Generation with Entropy-Guided Prioritized Progressive Learning},
  author={Li, Changlin and Zhang, Jiawei and Liu, Shuhao and Lin, Sihao and Shi, Zeyi and Li, Zhihui and Chang, Xiaojun},
  journal={arXiv preprint arXiv:2511.21136},
  year={2025}
}

@article{guo2025demospeedup,
  title={DemoSpeedup: Accelerating Visuomotor Policies via Entropy-Guided Demonstration Acceleration},
  author={Guo, Lingxiao and Xue, Zhengrong and Xu, Zijing and Xu, Huazhe},
  journal={arXiv preprint arXiv:2506.05064},
  year={2025}
}

@article{ben2025accelerated,
  title={Accelerated Sampling from Masked Diffusion Models via Entropy Bounded Unmasking},
  author={Ben-Hamu, Heli and Gat, Itai and Severo, Daniel and Nolte, Niklas and Karrer, Brian},
  journal={arXiv preprint arXiv:2505.24857},
  year={2025}
}

@article{zhang2025markovian,
  title={Markovian Scale Prediction: A New Era of Visual Autoregressive Generation},
  author={Zhang, Yu and Liu, Jingyi and Shi, Yiwei and Zhang, Qi and Miao, Duoqian and Wang, Changwei and Cao, Longbing},
  journal={arXiv preprint arXiv:2511.23334},
  year={2025}
}

@article{zhang2025actvar,
  title={ActVAR: Activating Mixtures of Weights and Tokens for Efficient Visual Autoregressive Generation},
  author={Zhang, Kaixin and Yang, Ruiqing and Zhang, Yuan and You, Shan and Huang, Tao},
  journal={arXiv preprint arXiv:2511.12893},
  year={2025}
}

@article{xie2024litevar,
  title={Litevar: Compressing visual autoregressive modelling with efficient attention and quantization},
  author={Xie, Rui and Zhao, Tianchen and Yuan, Zhihang and Wan, Rui and Gao, Wenxi and Zhu, Zhenhua and Ning, Xuefei and Wang, Yu},
  journal={arXiv preprint arXiv:2411.17178},
  year={2024}
}

@article{lifreqexit,
  title={Freqexit: Enabling early-exit inference for visual autoregressive models via frequency-aware guidance},
  author={Li, Ying and Lyu, Chengfei and Wang, Huan},
  journal={Advances in Neural Information Processing Systems},
  volume={38},
  pages={103763--103788},
  year={2026}
}

@article{qin2025head,
  title={Head-aware kv cache compression for efficient visual autoregressive modeling},
  author={Qin, Ziran and Lv, Youru and Lin, Mingbao and Guo, Hang and Zhang, Zeren and Zou, Danping and Lin, Weiyao},
  journal={arXiv preprint arXiv:2504.09261},
  year={2025}
}

@article{li2025memory,
  title={Memory-Efficient Visual Autoregressive Modeling with Scale-Aware KV Cache Compression},
  author={Li, Kunjun and Chen, Zigeng and Yang, Cheng-Yen and Hwang, Jenq-Neng},
  journal={arXiv preprint arXiv:2505.19602},
  year={2025}
}

@article{guo2025fastvar,
  title={Fastvar: Linear visual autoregressive modeling via cached token pruning},
  author={Guo, Hang and Li, Yawei and Zhang, Taolin and Wang, Jiangshan and Dai, Tao and Xia, Shu-Tao and Benini, Luca},
  journal={arXiv preprint arXiv:2503.23367},
  year={2025}
}

@inproceedings{chen2025frequency,
  title={Frequency-aware autoregressive modeling for efficient high-resolution image synthesis},
  author={Chen, Zhuokun and Fan, Jugang and Yu, Zhuowei and Zhuang, Bohan and Tan, Mingkui},
  booktitle={Proceedings of the IEEE/CVF International Conference on Computer Vision},
  pages={17140--17149},
  year={2025}
}

@article{li2025stagevar,
  title={StageVAR: Stage-Aware Acceleration for Visual Autoregressive Models},
  author={Li, Senmao and Wang, Kai and Khan, Salman and Khan, Fahad Shahbaz and Yang, Jian and Wang, Yaxing},
  journal={arXiv preprint arXiv:2512.16483},
  year={2025}
}

@article{li2025skipvar,
  title={SkipVAR: Accelerating Visual Autoregressive Modeling via Adaptive Frequency-Aware Skipping},
  author={Li, Jiajun and Ma, Yue and Zhang, Xinyu and Wei, Qingyan and Liu, Songhua and Zhang, Linfeng},
  journal={arXiv preprint arXiv:2506.08908},
  year={2025}
}

@article{xu2023imagereward,
  title={Imagereward: Learning and evaluating human preferences for text-to-image generation},
  author={Xu, Jiazheng and Liu, Xiao and Wu, Yuchen and Tong, Yuxuan and Li, Qinkai and Ding, Ming and Tang, Jie and Dong, Yuxiao},
  journal={Advances in Neural Information Processing Systems},
  volume={36},
  pages={15903--15935},
  year={2023}
}

@article{wu2023human,
  title={Human preference score v2: A solid benchmark for evaluating human preferences of text-to-image synthesis},
  author={Wu, Xiaoshi and Hao, Yiming and Sun, Keqiang and Chen, Yixiong and Zhu, Feng and Zhao, Rui and Li, Hongsheng},
  journal={arXiv preprint arXiv:2306.09341},
  year={2023}
}

@article{li2024playground,
  title={Playground v2. 5: Three insights towards enhancing aesthetic quality in text-to-image generation},
  author={Li, Daiqing and Kamko, Aleks and Akhgari, Ehsan and Sabet, Ali and Xu, Linmiao and Doshi, Suhail},
  journal={arXiv preprint arXiv:2402.17245},
  year={2024}
}

@article{dpgbench,
  title={Ella: Equip diffusion models with llm for enhanced semantic alignment},
  author={Hu, Xiwei and Wang, Rui and Fang, Yixiao and Fu, Bin and Cheng, Pei and Yu, Gang},
  journal={arXiv preprint arXiv:2403.05135},
  year={2024}
}

@article{ghosh2023geneval,
  title={Geneval: An object-focused framework for evaluating text-to-image alignment},
  author={Ghosh, Dhruba and Hajishirzi, Hannaneh and Schmidt, Ludwig},
  journal={Advances in Neural Information Processing Systems},
  volume={36},
  pages={52132--52152},
  year={2023}
}

@article{zhang2024var,
  title={Var-clip: Text-to-image generator with visual auto-regressive modeling},
  author={Zhang, Qian and Dai, Xiangzi and Yang, Ninghua and An, Xiang and Feng, Ziyong and Ren, Xingyu},
  journal={arXiv preprint arXiv:2408.01181},
  year={2024}
}

@inproceedings{han2025infinity,
  title={Infinity: Scaling bitwise autoregressive modeling for high-resolution image synthesis},
  author={Han, Jian and Liu, Jinlai and Jiang, Yi and Yan, Bin and Zhang, Yuqi and Yuan, Zehuan and Peng, Bingyue and Liu, Xiaobing},
  booktitle={Proceedings of the Computer Vision and Pattern Recognition Conference},
  pages={15733--15744},
  year={2025}
}

@article{wang2025training,
  title={Training-Free Text-Guided Image Editing with Visual Autoregressive Model},
  author={Wang, Yufei and Guo, Lanqing and Li, Zhihao and Huang, Jiaxing and Wang, Pichao and Wen, Bihan and Wang, Jian},
  journal={arXiv preprint arXiv:2503.23897},
  year={2025}
}

@misc{wu2025nestedautoregressivemodels,
      title={Nested AutoRegressive Models}, 
      author={Hongyu Wu and Xuhui Fan and Zhangkai Wu and Longbing Cao},
      year={2025},
      eprint={2510.23028},
      archivePrefix={arXiv},
      primaryClass={cs.CV},
      url={https://arxiv.org/abs/2510.23028}, 
}

@article{jiao2025flexvarflexiblevisualautoregressive,
  title={Flexvar: Flexible visual autoregressive modeling without residual prediction},
  author={Jiao, Siyu and Zhang, Gengwei and Qian, Yinlong and Huang, Jiancheng and Zhao, Yao and Shi, Humphrey and Ma, Lin and Wei, Yunchao and Jie, Zequn},
  journal={arXiv preprint arXiv:2502.20313},
  year={2025}
}

@article{qu2025visual,
  title={Visual autoregressive modeling for image super-resolution},
  author={Qu, Yunpeng and Yuan, Kun and Hao, Jinhua and Zhao, Kai and Xie, Qizhi and Sun, Ming and Zhou, Chao},
  journal={arXiv preprint arXiv:2501.18993},
  year={2025}
}

@article{rajagopalan2025restorevar,
  title={RestoreVAR: Visual Autoregressive Generation for All-in-One Image Restoration},
  author={Rajagopalan, Sudarshan and Narayan, Kartik and Patel, Vishal M},
  journal={arXiv preprint arXiv:2505.18047},
  year={2025}
}

@article{li2024controlvar,
  title={Controlvar: Exploring controllable visual autoregressive modeling},
  author={Li, Xiang and Qiu, Kai and Chen, Hao and Kuen, Jason and Lin, Zhe and Singh, Rita and Raj, Bhiksha},
  journal={arXiv preprint arXiv:2406.09750},
  year={2024}
}

@article{meni2024entropy,
  title={Entropy-based guidance of deep neural networks for accelerated convergence and improved performance},
  author={Meni, Mackenzie J and White, Ryan T and Mayo, Michael L and Pilkiewicz, Kevin R},
  journal={Information Sciences},
  volume={681},
  pages={121239},
  year={2024},
  publisher={Elsevier}
}

@article{peer2022improving,
  title={Improving the trainability of deep neural networks through layerwise batch-entropy regularization},
  author={Peer, David and Keulen, Bart and Stabinger, Sebastian and Piater, Justus and Rodr{\'\i}guez-S{\'a}nchez, Antonio},
  journal={arXiv preprint arXiv:2208.01134},
  year={2022}
}

@inproceedings{dubey2017regularizing,
  title={Regularizing prediction entropy enhances deep learning with limited data},
  author={Dubey, Abhimanyu and Gupta, Otkrist and Raskar, Ramesh and Rahwan, Iyad and Naik, Nikhil},
  booktitle={Proceedings of the Neural Information Processing Systems (NIPS)},
  year={2017}
}
\bibliographystyle{icml2026}

\end{document}